\documentclass[letterpaper]{article} 
\usepackage{aaai25}  
\usepackage{times}  
\usepackage{subfigure}
\usepackage{helvet}  
\usepackage{courier}  
\usepackage[hyphens]{url}  
\usepackage{graphicx} 
\usepackage{natbib}  
\usepackage{caption}
\usepackage{tabularx}
\usepackage{xcolor}
\usepackage{amsmath}
\usepackage{graphicx}
\usepackage{algorithm}
\usepackage{algorithmic}

\usepackage{booktabs,caption}
\usepackage[flushleft]{threeparttable}
\usepackage{newfloat}
\usepackage{listings}
\DeclareCaptionStyle{ruled}{labelfont=normalfont,labelsep=colon,strut=off} 
\floatstyle{ruled}
\newfloat{listing}{tb}{lst}{}
\floatname{listing}{Listing}
{\lccode`\?=`\p \lccode`\!=`\t  \lowercase{\gdef\ignorept#1?!{#1}}}
\def\divbyccvv#1{\expandafter\ignorept\the\dimexpr#1pt/255\relax}

\def\defineCMYKcolor#1#2{\defineCMYKcolorA{#1}#2,} 
\def\defineCMYKcolorA#1#2,#3,#4,#5,{\edef\tmp{\noexpand\definecolor{#1}{cmyk}%
      {\divbyccvv{#2},\divbyccvv{#3},\divbyccvv{#4},\divbyccvv{#5}}}\tmp
}

\defineCMYKcolor{my_color_name}{255,0,50,11}

\defineCMYKcolor{blue}{87,75,0,25}
\defineCMYKcolor{red}{0,77,77,43}
\defineCMYKcolor{green}{71,0,59,60}
\defineCMYKcolor{orange}{0,59,94,32}
\defineCMYKcolor{purple}{0,63,6,57}

\title{Hateful Meme Detection through Context-Sensitive Prompting\\ and Fine-Grained Labeling}
\author {
  Rongxin Ouyang,\textsuperscript{\rm 1} Kokil Jaidka,\textsuperscript{\rm 1,2} Subhayan Mukerjee,\textsuperscript{\rm 1,2} Guangyu Cui\textsuperscript{\rm 2}
}
\affiliations {
  \textsuperscript{\rm 1} Department of Communications and New Media, National University of Singapore\\
  \textsuperscript{\rm 2} Centre for Trusted Internet \& Community, National University of Singapore\\
  rongxin@u.nus.edu, jaidka@nus.edu.sg, mukerjee@nus.edu.sg, cui\_gy@nus.edu.sg
}
\usepackage{bibentry}
\begin{document}
\maketitle
\begin{abstract}
  The prevalence of multi-modal content on social media complicates automated moderation strategies. This calls for an enhancement in multi-modal classification and a deeper understanding of understated meanings in images and memes. Although previous efforts have aimed at improving model performance through fine-tuning, few have explored an end-to-end optimization pipeline that accounts for modalities, prompting, labeling, and fine-tuning. In this study, we propose an end-to-end conceptual framework for model optimization in complex tasks. Experiments support the efficacy of this traditional yet novel framework, achieving the highest accuracy and AUROC. Ablation experiments demonstrate that isolated optimizations are not ineffective on their own.
\end{abstract}

\begin{links}
  \link{Code}{https://github.com/reycn/multi-modal-scale}
  \link{Datasets}{https://ai.meta.com/blog/hateful-memes-challenge-and-data-set/}
\end{links}

\section{Introduction}

Recent years have seen a significant increase in visual content on social media~\cite{peng2023agenda, heley2022missing}, particularly visual misinformation~\cite{yang2023visual}. Up to 30\% of the content on platforms like X includes images or videos~\cite{pfeffer2023just}, highlighting the need for a multi-modal research on social media.

However, while increasingly more studies have recognized the visual moderation challenge~\cite{gonzalez2023populist,solea2023mainstreaming}, most prior work has either unimodal~\cite{muddiman2019re}, or focused on fine-tuning only~\cite{lippe2020multimodal, hermida2023detecting}. Prior work on prompt engineering~\cite{furniturewala2024thinking} indicates the relative advantage of multi-stage prompts that act to pre-empt biases in Large Language Model (LLM) outputs. Yet, these studies focus on unimodal content, and it is unclear whether using multi-stage prompts suffices to improve the classification accuracy in Vision Language Model (VLM) outputs for multimodal input. In the case of Large- and Multimodal Language Models, it is unclear whether fine-grained categories for prompting and labeling outweigh the effect of fine-tuning. To address these research gaps, our experiment design systematically evaluates the contributions of each factor to determine whether combining them enhances performance in multimodal hate speech detection, with applications for content moderation.

\begin{figure}
  \centering
  \includegraphics[width=0.6\linewidth]{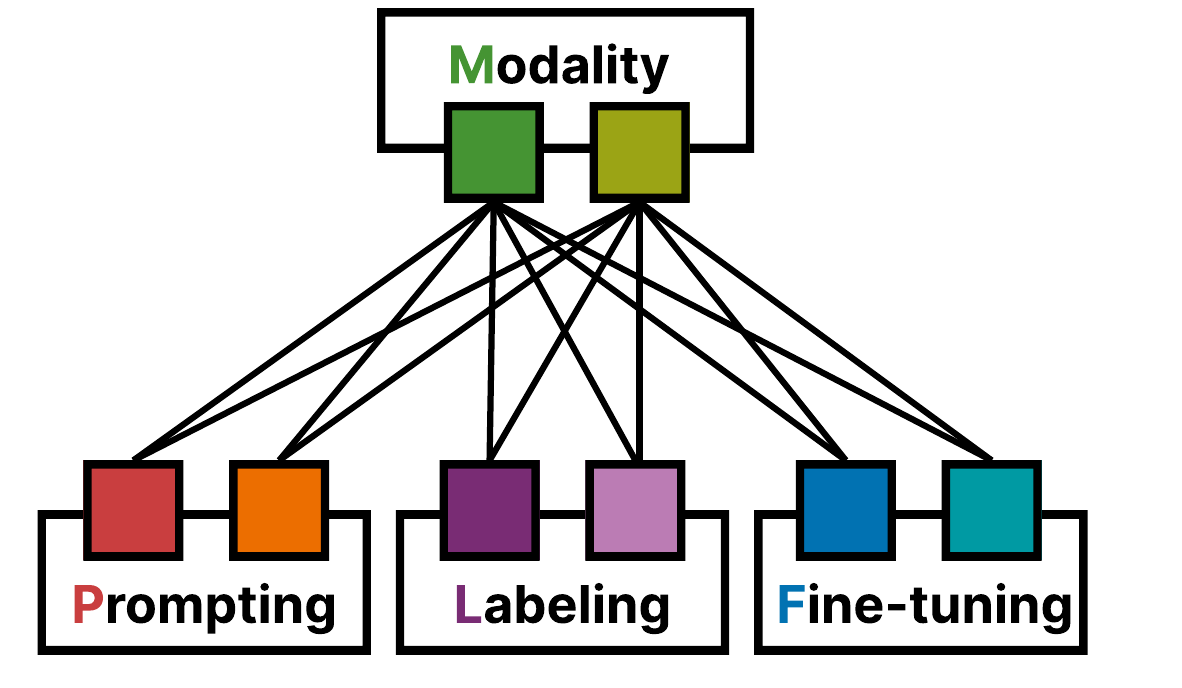}
  \caption{A Conceptual Framework}
  \label{fig:framework}
\end{figure}

\begin{figure*}[!htp]
  \centering
  \subfigure[Accuracy]{
    \includegraphics[width=0.45\linewidth]{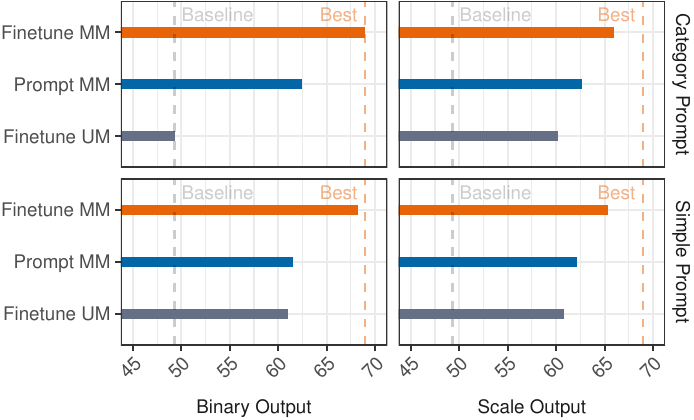}
  }\label{fig:accuracy}
  \subfigure[AUROC]{
    \includegraphics[width=0.376\linewidth]{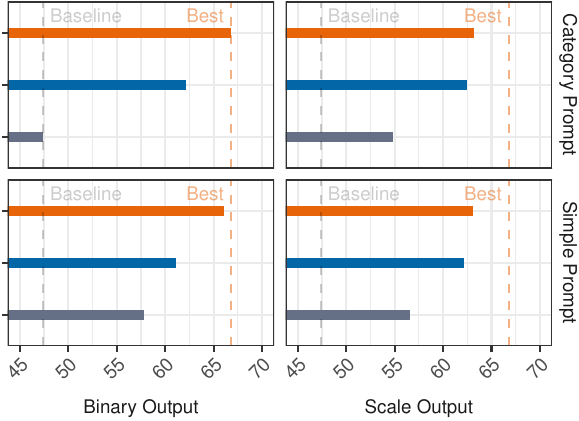}
  }\label{fig:auroc}
  \caption{Model Performance of Combinations (\%, MM=multi-modal, UM=unimodal)}
\end{figure*}

\begin{table*}[!t]
\setlength{\tabcolsep}{1.53mm}
  \centering
    \begin{tabular}{lllllrrrrr}
      \hline
      \textbf{Category} & \textbf{Model}  & \textbf{Size} & \textbf{Prompt} & \textbf{Label} & \textbf{Accuracy} & \textbf{Precision} & \textbf{Recall} & \textbf{F1-score} & \textbf{AUROC} \\
      \hline
      Prompting  & InternVL & 8B & Simple$^{(a)}$ & Binary$^{(c)}$ &61.500 & \underline{68.041}$^{*}$ & 40.408 & 50.704 & 61.086  \\

         \textit{(multi-modal)} & & &  & Scale$^{(d)}$ &62.100 & 60.491 &
    \underline{65.306} & \underline{62.807} & 62.163 \\

      &  &  & Category$^{(b)}$& Binary &62.500 & 66.571 & 47.143 & 55.197 & 62.199 \\
      &  & &  & Scale &62.700& 66.387 & 48.367 & 55.962 & \underline{62.419}\\
      \hline

      Fine-tuning  & InternVL & 8B & Simple & Binary &68.233 &63.811 &53.468 &58.183 & 66.052 \\
      \textit{(multi-modal)} &  & &  & Scale &65.367 &59.673 &50.000 &54.410 & 63.097 \\
      &  &  & Category & Binary (\textbf{M})&\underline{68.933} &64.695 &54.677 &59.266 & \underline{66.827} \\
      &  & &  & Scale &65.933 &61.498 &47.016 &53.291 & 63.139 \\
      \hline
      Fine-tuning  & Distil-Bert & 66M & Simple & Binary &61.033 &53.958 &39.032 &45.297 & 57.783 \\
      \textit{(unimodal)} &  & &  & Scale &60.833 &58.668 &56.612 &55.657 & 56.612 \\
      \ &  &  & Category$^{\#}$ & Binary &49.300 &38.103 &36.290 &37.174 & 47.378 \\
      &  & &  & Scale &60.167 &57.818 &54.793 &36.975 & 54.793 \\
      \hline
  \end{tabular}
  \begin{tablenotes}
\item Note. $^*$ Best results are underlined. $^\#$ The loss was decreasing slowly, but we maintained the same parameters for comparison.
\end{tablenotes}
\caption{Experimental Results (\%)}
\label{tab:result}
\end{table*}

\section{Methodology}
The principle underlying the proposed framework is captured by Equation 1: we consider performance ($\delta$) as a multivariate optimization problem dependent on modalities ($M$), prompting ($P$), labeling ($L$), and fine-tuning ($F$). 
\begin{equation}
\begin{split}
\delta = f(\textcolor{green}{M},\textcolor{red}{P},\textcolor{purple}{L},\textcolor{blue}{F})
\end{split}
\end{equation}

More specifically, we first compare modalities, between a visual model, InternVL 2~\cite{chen2023internvl} (8B), with another text-based model, DistilBERT~\cite{sanh2020distilbert} (66M); and expect a better performance of the multi-modal approach for more information recognized. Second, we use both prompting and fine-tuning on the same model, InternVL 8B, with identical prompting and labeling strategies. Last we compiled a $2\times 2$ matrix of construct by both prompting (simple question or categories defined in details) and labeling (binary output or outputs in an interval scale). A simple prompt asks a plain question while categories provide detailed definitions of sub-categories of hateful content (see SI). To get labels in scales for fine-tuning, we used GPT4-o-mini to generate answers in scales and excluded incorrectly annotated cases in training according to the ground truth, ensuring the quality of the extended annotations. More details about prompts hyper-parameters are documented in Supplementary Information.

Based on combinations of those strategies, we conducted 12 experiments with a $3 \times 2\times 2$ ablation design. These combinations of settings (see Table 1) ablate modalities (unimodal or multi-modal), prompting strategies (simple or category), labeling strategies (binary or scaled outputs) and fine-tuning process. To supplement the ground-truth of scales for fine-tuning, we used GPT-4o-mini and selected the correct ones. More details are provided in SI.

We used the Facebook Hateful Memes dataset~\cite{kiela2020hateful} for experiments. It includes more than 10k images, human captions and binary labels of hatefulness for training; as well as 3k entries for evaluation. Performance were evaluated  by ACCU and AUROC.

\section{Results and Future Work}
As shown in Table~\ref{tab:result}, the best model is not the one with highest complexity, highlighting the necessity of our framework. Among all the models, the model \textbf{M} achieves the highest accuracy (68.933\%, +19.611 p.p.) and AUROC (66.827\%, +19.449 p.p.). Comparisons of ablation show that this improvement results from fine-tuning, categorical prompting, and binary labels. However, components beneficial to model \textbf{M} do not universally enhance performance (e.g., scaled outputs generally improve performance but not always), underscoring the need for the end-to-end framework we proposed.

In summary, our study introduces an end-to-end optimization pipeline for complex, multi-modal tasks like hateful meme detection. Our experiments demonstrate the effectiveness of a global optimization strategy within this framework. Moreover, our ablation studies indicate that isolated optimizations are not better by themselves (e.g., scales improve performance in most settings but not on the best model). We therefore argue that this traditional wisdom is both beneficial and necessary for such complicated, modern tasks.

\pagebreak
\section*{Acknowledgments}
This work was supported by the Singapore Ministry of Education AcRF TIER 3 Grant (MOE-MOET32022-0001). We gratefully acknowledge invaluable comments and discussions with Shaz Furniturewala and Jingwei Gao.

\section*{Supplementary Material}

This supplementary material includes detailed prompts (Table 1), experimental settings (section 2), and simplified core codes for both training and evaluations (section 3).

\section{Prompting Components}
All the prompting strategies in this paper are divided into specific modules (see Table 1). The ``simple'' prompting component asks the model a straightforward question. The ``category'' component breaks down the question into specific subcategories of hatefulness. The ``scale'' labeling component requires numerical outputs (e.g., 0-9), while the ``binary'' component expects a boolean value. To minimize conversion errors, we implemented output constraints to either the ``scale'' or ``binary'' group. To control confounders, we restrict the prompts to clean combinations of several components, even though some other strategies perform better. All of them were developed based on prior works~\cite{furniturewala2024thinking}.

\begin{table*}[]
  \centering
  \begin{tabularx}{\textwidth}{X p{12.9cm}}
    \hline
    \textbf{Dataset} & \textbf{Prompt Example} \\
    \hline
    \textcolor{blue}{\{\{Simple\}\}} + \textcolor{gray}{\{\{Binary\}\}}   &  \textcolor{blue}{\{\{Your task is to analyze this given image and its caption to identify if there's any forms of hateful content.\}\}} \textcolor{gray}{\{\{If it's hateful, return the `TRUE` | `FALSE`.\}\}}\\
    \hline

    \textcolor{blue}{\{\{Simple\}\}} + \textcolor{green}{\{\{Scale\}\}} &  \textcolor{blue}{\{\{Simple\}\}}
    \textcolor{green}{\{\{You will be provided with the image, based on this label and the content of the image, you will assign a
          hatefulness score on a scale from 0 to 9, where:
          0\: The image is not hateful at all.9\: The image is extremely hateful.
    Return the score as an integer in range 0 to 9.\}\}}\\
    \hline
    \textcolor{orange}{\{\{Category\}\}}  + \textcolor{gray}{\{\{Binary\}\}}&
    \textcolor{blue}{\{\{Simple\}\}}\textcolor{orange}{\{\{Try to focus on the presence of any element that relates to any of the following:
          1. Sexual aggression:
          a. Homophobia and Transphobia: This category encompasses hate speech targeting LGBTQ+ individuals, including slurs,
          derogatory comments, and rhetoric that seeks to undermine or dehumanize people based on their sexual orientation or
          gender identity.
          b. Misogyny and Sexism: This category includes hate speech directed at women or based on gender. It covers
          derogatory language, stereotypes, and rhetoric that perpetuate gender inequality, objectification, and violence
          against women.
          2. Hate based on ideology:
          a. Political Hate Speech: This category includes hate speech that is politically motivated, often targeting
          individuals or groups based on their political beliefs or affiliations. It may include inflammatory language,
          threats, and rhetoric designed to polarize or incite violence within political contexts.
          3. Racism and xenophobia:
          a. COVID-19 and Xenophobia: This category includes hate speech that arose during the COVID-19 pandemic, often
          targeting specific ethnic groups or nationalities. It includes xenophobic language blaming certain groups for the
          spread of the virus, as well as fear-mongering and scapegoating related to the pandemic.
          b. Racism Against Black People: This category focuses on hate speech directed at Black individuals or communities.
          It includes racial slurs, stereotypes, dehumanization, and other forms of derogatory language that perpetuate
          racial discrimination and inequality.
          c. Racist Hate Against Other Ethnic Groups: This category includes hate speech directed at various ethnic groups
          other than Black individuals. It covers a range of racial slurs, xenophobic language, dehumanization, and
          derogatory remarks targeting specific ethnicities or nationalities.
          d. White Supremacy: This category includes hate speech promoting white supremacist ideology, often intertwined with
          Christian extremist views. It includes rhetoric that advocates for racial superiority, anti-immigrant sentiments,
          and the dehumanization of non-white groups, sometimes using religious justifications for these beliefs.
          4. Bigotry:
          a. Anti-Muslim and Islamophobic Hate: This category comprises hate speech aimed at Muslims and Islam. It includes
          language that promotes fear, hatred, dehumanization, or prejudice against Muslim individuals or communities, often
          using religious and cultural references to incite hostility.
          b. Anti-Semitic Hate: This category focuses on hate speech directed at Jewish people and Judaism. It includes
          references to historical anti-Semitic tropes, conspiracy theories, and other forms of rhetoric that seek to
          dehumanize or discriminate against Jewish individuals and communities.
          5. Miscellaneous Hate Speech: This category captures various forms of hate speech that do not fit neatly into the
          other categories. It includes a wide range of derogatory language and expressions that target individuals or groups
          based on different aspects of identity or personal characteristics. This category includes hate speech that targets
          individuals based on their physical or mental disabilities. It often includes derogatory language that mocks or
    devalues people with disabilities, promoting harmful stereotypes and exclusion.\}\}} \textcolor{gray}{\{\{Binary\}\}}\\

    \hline
    \textcolor{orange}{\{\{Category\}\}} + \textcolor{green}{\{\{Scale\}\}}  &
    \textcolor{blue}{\{\{Simple\}\}}\textcolor{orange}{\{\{Category\}\}}\textcolor{green}{\{\{Scale\}\}}\\
    \hline
  \end{tabularx}
  \begin{tablenotes}
\item Note. a) Brackets are only used for clarifying; b) each color represents a unique component of prompts.
\end{tablenotes}
\caption{Prompts used for each experiment}
\label{tab:prompt}
\end{table*}

\section{Detailed Settings}
The experiments includes three dimensions. The first dimension is model category: multi-modal prompting on a large pre-trained model, \textit{InternVL}~\cite{chen2023internvl}, multi-modal fine-tuning of the same large pre-trained model, and a unimodal fine-tuning on a smaller textual model, \textit{DistilBERT}~\cite{sanh2020distilbert}. To effectively simulate the common computation capacity, we used \textit{LoRA}~\cite{hu2021lora}, low-rank adaptations of large language models, to reduce computation costs in the fine-tuning of 8B models.

The second dimension is the prompting strategy: either a basic prompt for this classification task (prompt component \textit{simple}), or a detailed version defining potential categories of hatefulness (prompt component \textit{category}; see Table 1). Then we introduce the third dimension of output format: binary or scales. Binary label refers to a direct question about the result of the classification (i.e., True or False, label component \textit{binary}); on the other hand, scales potentially capture more fine-grained levels of hatefulness (label component \textit{scale}). For comparison, we keep the same settings for each component across all combinations.

Since the original dataset lacks scaled outputs for verifying the pipeline, we used GPT-4o-mini as a teacher model to generate labels. To ensure 100\% accuracy in this annotation process, we manually filtered out incorrect entries using the binary ground-truth. For example, if the ground-truth is hateful but the model rates it 1 (not hateful) on a 0-9 scale, we removed it. This process not only produced accurate outputs but also provided additional information from multi-modal representations to scales from the teacher model.

Hyper-parameters were defined as follows. For InternVL, we used the default hyper-parameters for fine-tuning. For DistilBERT, we applied the same settings listed below to all models. These decisions were intended to control confounders.
\\

\lstset{basicstyle=\ttfamily\color{blue}}
\begin{lstlisting}
...
training_args = TrainingArguments(
    output_dir="../process/bert-baseline",
    learning_rate=2e-5,
    per_device_train_batch_size=12,
    per_device_eval_batch_size=12,
    num_train_epochs=12,
    weight_decay=0.01,
    eval_strategy="epoch",
    save_strategy="epoch",
    load_best_model_at_end=True,
)
...
\end{lstlisting}

\section{Simplified Codes of Training and Evaluation}
Here, we illustrate the automatic, batched fine-tuning and evaluation process with a simplified structure. Full code is available on GitHub.

\lstset{basicstyle=\ttfamily\color{blue}}
\begin{lstlisting}
...
define_variables_and_functions()
for model in models:
    for prompt in prompts:
        train_set, test_set = import_datasets().split()
        train_set, test_set = tokenize(train_set), tokenize(test_set)
        model = import_model(model)
        finetune(prompt, model, train_set)
        predictions = predict(test_set)
        merge(test_set, predictions)
        convert_predictions() # if needed, e.g., scales
        compute_metrics(["accuracy", "precision", "recall", "f1", "auroc"]).save()
...
\end{lstlisting}
\bibliography{aaai25}

\begin{thebibliography}{14}
\providecommand{\natexlab}[1]{#1}

\bibitem[{Chen et~al.(2023)Chen, Wu, Wang, Su, Chen, Xing, Zhong, Zhang, Zhu, Lu, Li, Luo, Lu, Qiao, and Dai}]{chen2023internvl}
Chen, Z.; Wu, J.; Wang, W.; Su, W.; Chen, G.; Xing, S.; Zhong, M.; Zhang, Q.; Zhu, X.; Lu, L.; Li, B.; Luo, P.; Lu, T.; Qiao, Y.; and Dai, J. 2023.
\newblock InternVL: Scaling up Vision Foundation Models and Aligning for Generic Visual-Linguistic Tasks.
\newblock \emph{arXiv preprint arXiv:2312.14238}.

\bibitem[{Furniturewala et~al.(2024)Furniturewala, Jandial, Java, Banerjee, Shahid, Bhatia, and Jaidka}]{furniturewala2024thinking}
Furniturewala, S.; Jandial, S.; Java, A.; Banerjee, P.; Shahid, S.; Bhatia, S.; and Jaidka, K. 2024.
\newblock Thinking Fair and Slow: On the Efficacy of Structured Prompts for Debiasing Language Models.
\newblock \emph{arXiv preprint arXiv:2405.10431}.

\bibitem[{Gonz{\'a}lez-Aguilar, Segado-Boj, and Makhortykh(2023)}]{gonzalez2023populist}
Gonz{\'a}lez-Aguilar, J.~M.; Segado-Boj, F.; and Makhortykh, M. 2023.
\newblock Populist Right Parties on TikTok: Spectacularization, Personalization, and Hate Speech.
\newblock \emph{Media and communication}, 11(2): 232--240.

\bibitem[{Heley, Gaysynsky, and King(2022)}]{heley2022missing}
Heley, K.; Gaysynsky, A.; and King, A.~J. 2022.
\newblock Missing the bigger picture: The need for more research on visual health misinformation.
\newblock \emph{Science communication}, 44(4): 514--527.

\bibitem[{Hermida and Santos(2023)}]{hermida2023detecting}
Hermida, P. C. D.~Q.; and Santos, E. M.~D. 2023.
\newblock Detecting Hate Speech in Memes: A Review.
\newblock \emph{Artificial Intelligence Review}, 56(11): 12833--12851.

\bibitem[{Hu et~al.(2021)Hu, Shen, Wallis, {Allen-Zhu}, Li, Wang, Wang, and Chen}]{hu2021lora}
Hu, E.~J.; Shen, Y.; Wallis, P.; {Allen-Zhu}, Z.; Li, Y.; Wang, S.; Wang, L.; and Chen, W. 2021.
\newblock {{LoRA}}: Low-Rank Adaptation of Large Language Models.
\newblock arXiv:2106.09685.

\bibitem[{Kiela et~al.(2020)Kiela, Firooz, Mohan, Goswami, Singh, Ringshia, and Testuggine}]{kiela2020hateful}
Kiela, D.; Firooz, H.; Mohan, A.; Goswami, V.; Singh, A.; Ringshia, P.; and Testuggine, D. 2020.
\newblock The hateful memes challenge: Detecting hate speech in multimodal memes.
\newblock \emph{Advances in neural information processing systems}, 33: 2611--2624.

\bibitem[{Lippe et~al.(2020)Lippe, Holla, Chandra, Rajamanickam, Antoniou, Shutova, and Yannakoudakis}]{lippe2020multimodal}
Lippe, P.; Holla, N.; Chandra, S.; Rajamanickam, S.; Antoniou, G.; Shutova, E.; and Yannakoudakis, H. 2020.
\newblock A Multimodal Framework for the Detection of Hateful Memes.
\newblock arXiv:2012.12871.

\bibitem[{Muddiman, McGregor, and Stroud(2019)}]{muddiman2019re}
Muddiman, A.; McGregor, S.~C.; and Stroud, N.~J. 2019.
\newblock (Re) claiming our expertise: Parsing large text corpora with manually validated and organic dictionaries.
\newblock \emph{Political Communication}, 36(2): 214--226.

\bibitem[{Peng, Lu, and Shen(2023)}]{peng2023agenda}
Peng, Y.; Lu, Y.; and Shen, C. 2023.
\newblock An Agenda for Studying Credibility Perceptions of Visual Misinformation.
\newblock \emph{Political Communication}, 40(2): 225--237.

\bibitem[{Pfeffer et~al.(2023)Pfeffer, Matter, Jaidka, Varol, Mashhadi, Lasser, Assenmacher, Wu, Yang, Brantner et~al.}]{pfeffer2023just}
Pfeffer, J.; Matter, D.; Jaidka, K.; Varol, O.; Mashhadi, A.; Lasser, J.; Assenmacher, D.; Wu, S.; Yang, D.; Brantner, C.; et~al. 2023.
\newblock Just another day on Twitter: a complete 24 hours of Twitter data.
\newblock In \emph{Proceedings of the International AAAI Conference on Web and Social Media}, volume~17, 1073--1081.

\bibitem[{Sanh et~al.(2020)Sanh, Debut, Chaumond, and Wolf}]{sanh2020distilbert}
Sanh, V.; Debut, L.; Chaumond, J.; and Wolf, T. 2020.
\newblock {{DistilBERT}}, a Distilled Version of {{BERT}}: Smaller, Faster, Cheaper and Lighter.
\newblock arXiv:1910.01108.

\bibitem[{Solea and Sugiura(2023)}]{solea2023mainstreaming}
Solea, A.~I.; and Sugiura, L. 2023.
\newblock Mainstreaming the Blackpill: Understanding the Incel Community on TikTok.
\newblock \emph{European Journal on Criminal Policy and Research}, 1--26.

\bibitem[{Yang, Davis, and Hindman(2023)}]{yang2023visual}
Yang, Y.; Davis, T.; and Hindman, M. 2023.
\newblock Visual misinformation on Facebook.
\newblock \emph{Journal of Communication}, jqac051.

\end{thebibliography}
\end{document}